\documentclass{article} 
\usepackage{arxiv}

\usepackage{amsmath,amsfonts,bm}

\def\eqref#1{equation~\ref{#1}}

\def\1{\bm{1}}

\DeclareMathAlphabet{\mathsfit}{\encodingdefault}{\sfdefault}{m}{sl}
\SetMathAlphabet{\mathsfit}{bold}{\encodingdefault}{\sfdefault}{bx}{n}

\usepackage{natbib}
\usepackage{multirow}
\usepackage{hyperref}
\usepackage{url}
\usepackage{mathtools}
\usepackage{amsthm}

\usepackage{amssymb}
\usepackage{algorithm}
\usepackage{algorithmic}
\usepackage{adjustbox}

\usepackage{amsmath}
\usepackage{booktabs}
\usepackage{siunitx,tabularx,adjustbox,xcolor}
\newcommand{\TopK}{\operatorname{TopK}}
\newtheorem{thm}{Theorem}[section] 

\newtheorem{cor}[thm]{Corollary}

\title{DynaSpec: Context-aware Dynamic Speculative Sampling for Large-Vocabulary Language Models}

\author{Jinbin Zhang \\
Department of Computer Science\\
Aalto University\\
Espoo, Finland \\
\texttt{jinbin.zhang@aalto.fi} \\
\And
Nasib Ullah \\
Department of Computer Science \\
Aalto University\\
Espoo, Finland \\
\texttt{nasibullah.nasibullah@aalto.fi} \\
\And
Erik Schultheis \\
 Deep Algorithms and Systems Lab  \\
IST Austria\\
Austria \\
\texttt{erik.schultheis@ist.ac.at} \\
\And
Rohit Babbar \\
Department of Computer Science \\
University of Bath\\
United Kingdom\\
\texttt{rb2608@bath.ac.uk} \\
}

\newcommand{\Comment}[1]{\hfill\texttt{//}~#1}

\begin{document}

\maketitle

\begin{abstract}
Speculative decoding accelerates LLM inference by letting a small drafter propose multiple tokens which a large target model verifies once per speculation step. As vocabularies scale past $10^5$ tokens, verification cost in the target model is largely unchanged, but the drafter can become bottlenecked by its $\mathcal{O}(|V|d)$ output projection. Recent approaches (e.g., FR-Spec, VocabTrim) mitigate this by restricting drafting to a fixed, frequency-ranked shortlist; however, such static truncation is corpus-dependent and suppresses rare or domain-specific tokens, reducing acceptance and limiting speedups. 
We propose DynaSpec, a context-dependent dynamic shortlisting mechanism for large-vocabulary speculative decoding. DynaSpec trains lightweight meta-classifiers that route each context to a small set of coarse token clusters; the union of the top-selected clusters defines the drafter’s shortlist, while the target model still verifies over the full vocabulary, preserving exactness. Systems-wise, routing is overlapped with draft computation via parallel execution streams, reducing end-to-end overhead. 
Across standard speculative decoding benchmarks, DynaSpec consistently improves mean accepted length—recovering 98.4\% of full-vocabulary performance for Llama-3-8B versus 93.6\% for fixed-shortlist baselines—and achieves up to a 2.23× throughput gain compared to 1.91× for static approaches on the dataset with rare tokens.
\end{abstract}

\section{Introduction}
Over the past few years, large language models (LLMs)~\citep{brown2020language,touvron2023llama,liu2024deepseek,jiang2024mixtral} have improved dramatically, reshaping products and workflows across sectors. Beyond algorithmic advances, much of this progress has come from scaling data and especially model parameters~\citep{kaplan2020scaling,hoffmann2022training}. Yet these gains come with a cost: inference becomes compute-heavy and latency-sensitive, which limits real-time applications~\citep{kwon2023efficient,yuan2024llm}. As agentic systems~\citep{yao2023react,wang2024survey} and reasoning-heavy workflows (e.g., chain-of-thought)~\citep{wei2022chain,zhang2024chain} proliferate, the demand for faster inference keeps rising. Speculative decoding~\citep{chen2023accelerating,leviathan2023fast,li2024eagle,li2024eagle2} is a practical way to accelerate generation. A smaller draft model proposes multiple tokens so the target model need not run at every step, while the rejection sampling procedure guarantees that the final output distribution matches that of the target model. In the worst case, quality and latency revert to standard decoding; in typical cases, throughput improves substantially.

Recently~\textit{scaling laws} for vocabulary~\citep{tao2024scaling,huangover,yu2025scaling} suggest that larger models benefit from larger vocabularies, and deployed LLMs have indeed expanded their vocabularies over time. For example, tokenizer vocabularies have scaled from 32k entries in Llama-2~\citep{touvron2023llama} to 128k in Llama-3~\citep{grattafiori2024llama}, 129k in DeepSeek-V3~\citep{liu2024deepseek}, 152k in Qwen-2.5~\citep{yang2025qwen3}, and 262k in Gemma-3~\citep{team2025gemma}.  
While the large vocabulary size contributes only marginally to the inference cost of the target model’s output layers given its overall model size, it imposes a substantial computational burden on the draft model during speculative decoding, as the draft model is much smaller and the output layer constitutes a larger fraction of its total computation, thereby diminishing the overall pipeline efficiency. To mitigate this, FR-Spec~\citep{zhao-etal-2025-fr} and Vocabtrim~\citep{goel2025vocabtrim} replace the full vocabulary with a fixed shortlist---typically the top-p\% of the most frequent tokens measured on a reference corpus. This speeds up drafting, but it is suboptimal for two reasons: (1) frequency lists are corpus-dependent and often fail to generalize across benchmarks, requiring substantial retuning; and (2) a static subset of the token vocabulary can degrade performance on tasks that rely on rare or domain-specific tokens, where maintaining diversity in the candidate set is critical for high-quality outputs.

In this work, we move beyond fixed shortlists and introduce a \emph{dynamic vocabulary head}. We first define \emph{meta-labels} as coarse token clusters: the vocabulary $V$ is partitioned into \(M\!\ll\!|V|\) clusters \(\{C_m\}_{m=1}^M\) by clustering column-normalized LM-head weights. Conditioned on current context (in the form of embedding and features), a lightweight \emph{meta-classifier} scores clusters and we evaluate the drafter’s LM head only on the union of the selected clusters \(V_S=\bigcup_{m\in\mathcal{K}}\!C_m\), while verification still runs over the full vocabulary, preserving exactness. This coarse-to-fine routing, inspired by \citet{jiang2021lightxml} and ~\citet{kharbanda2022cascadexml}, replaces an \(\mathcal{O}(|V|d)\) head with \(\mathcal{O}\!\big((M+|V_S|)d\big)\). Making the support context-dependent leads to a consistent increase (across datasets) in the mean accepted length relative to the static subset proposed in \citet{zhao-etal-2025-fr}. However, a dynamic (indexed) head introduces gathered matmul overhead and can be slower than a fixed, pre-trimmed dense head. We, therefore, couple the dynamic head with a \emph{position-aware} cluster budget: early tokens receive larger shortlists, and the budget decays with position. As argued in recent works, \cite{li2025gumiho,li2024eagle2}, prioritizing earlier tokens improves expected acceptance, while shrinking the shortlist later reduces latency, yielding net end-to-end gains without altering verification. Finally, we show the same position-aware schedule can retrofit frequency-ranked shortlists (FR-Spec)~\citep{zhao-etal-2025-fr}. 

\paragraph{Contributions} 
To address the drafter bottleneck in large-vocabulary speculative decoding, we introduce \textsc{DynaSpec}, a context-aware dynamic vocabulary head that trades a full $\mathcal{O}(|V|d)$ projection for a routed, sparse head while preserving exact verification. Specifically: 
\begin{itemize}
    \item \textit{Context-aware dynamic shortlisting via cluster routing}. We partition the vocabulary into coarse clusters and train a lightweight router to select a small set of clusters per decoding step; the drafter evaluates logits only on the union of selected clusters, while the target model continues to verify on the full vocabulary. 
    \item \textit{Theory: context-conditioned supports dominate static truncation}. We provide an acceptance analysis showing that, for a fixed budget, context-dependent supports can strictly improve expected overlap/acceptance relative to any fixed subset, and we relate router quality to an explicit ``routing regret'' term.
    \item \textit{Systems: amortizing routed-head overhead}. To reconcile dynamic shortlisting with the overhead of indexed (gathered) matmul, we (i) run routing on a parallel CUDA stream so its latency is largely hidden under draft computation, and (ii) introduce a position-aware cluster budget that allocates larger shortlists to early draft steps and decays thereafter; we further fuse index selection with the shortlisted-head GEMM in a custom kernel. 
    \item \textit{Empirical gains on large-vocabulary LLMs}. Across seven benchmarks, DynaSpec improves mean accepted length over static shortlists (e.g., 3.64 $\rightarrow$ 3.83 tokens/step on Llama-3-8B with a smaller $\sim$ 28K average shortlist vs. 32K), recovering 98.4\% of full-vocabulary acceptance versus 93.6\% for fixed-shortlist baselines, and yields up to a 2.23 $\times$ throughput gain (vs. 1.91$\times$ for static approaches) while remaining plug-in compatible with EAGLE-style pipelines.
\end{itemize}

\section{Related Work}

\subsection{LLMs with Large Vocabulary}
\label{sec:rw-vocab}
As LLMs have grown in scale, so too have their vocabularies. Early models such as GPT-3~\citep{brown2020language} and LLaMA-2~\citep{touvron2023llama} used vocabularies of $32\mathrm{k}$ tokens, while more recent releases exceed $100\mathrm{k}$ (e.g., $128\mathrm{k}$ in LLaMA-3~\citep{grattafiori2024llama}, $152\mathrm{k}$ in Qwen-2.5~\citep{yang2025qwen3}, and $262\mathrm{k}$ in Gemma-3~\citep{team2025gemma}). 
Large vocabularies are especially common in multilingual and high-coverage LLMs: mT5~\citep{xue2020mt5} adopts a $250\mathrm{k}$ SentencePiece~\citep{kudo2018sentencepiece} vocabulary to improve cross-lingual coverage, and PaLM uses a $256\mathrm{k}$ unigram vocabulary to shorten sequences and better encode code and symbols~\citep{chowdhery2023palm}. Recently, empirical scaling laws suggest that larger vocabularies improve expressivity and perplexity~\citep{tao2024scaling,yu2025scaling}, but they also magnify inference cost: the output projection grows linearly with vocabulary size, which can dominate latency in small drafters. 
\subsection{Speculative Decoding}
\label{sec:rw-speculative}
In line with Speculative Decoding, early work accelerates \emph{greedy} generation~\citep{stern2018blockwise,sun2021instantaneous}, and subsequent work extends the idea to \emph{non-greedy} sampling with distributional guarantees~\citep{leviathan2023fast,chen2023accelerating,xia2022speculative}. These methods differ primarily in how they form proposals: some reuse information already present in the prompt via retrieval-guided drafting---effective when outputs strongly overlap inputs~\citep{saxena2023prompt,yang2023inference,he2024rest}---whereas others learn parametric drafters for general-purpose speedups, ranging from parallel MLP heads (and serial variants) to lightweight transformer blocks that condition on richer context~\citep{cai2024medusa,ankner2024hydra}. The EAGLE series further refines this paradigm: \textsc{EAGLE} employs lightweight transformer drafters, while \textsc{EAGLE-2} introduces dynamic draft trees to boost efficiency~\citep{li2024eagle,li2024eagle2}. Orthogonal advances in training and systems—such as adaptive exploration, cache reuse, and efficient serving stacks—increase accepted tokens per verification and reduce overhead~\citep{kwon2023efficient,zheng2024sglang,dao2023flashattention}. Complementing these directions, \textsc{MagicDec} analyzes the latency–throughput trade-off and shows that speculative decoding remains effective at large batch sizes and with long contexts, whereas \textsc{TriForce} targets long-context regimes by compressing the drafter’s KV cache and coordinating reuse across draft steps~\citep{sadhukhan2024magicdec,sun2024triforce}. Most recently, \citet{timor2025accelerating} proposes lossless speculative decoding for \emph{heterogeneous vocabularies}, relaxing the same-tokenizer requirement between drafter and target; meanwhile, \citet{li2025gumiho} combines Medusa-style heads~\citep{cai2024medusa} with the EAGLE~\citep{li2024eagle2} framework. We target a different bottleneck: as vocabularies expand, the drafter’s output projection increasingly dominates latency.
\subsection{Accelerating Large Vocabularies}
\label{sec:rw-large-output}
The computational bottleneck of large output spaces or vocabularies has been recognized in both language modeling and neighboring fields.  
Within LLMs, FR-Spec~\citep{zhao-etal-2025-fr} and VocabTrim~\citep{goel2025vocabtrim} mitigate drafter overhead by pruning the output layer to a fixed shortlist of high-frequency tokens. While effective in reducing latency, these static subsets may suppress rare or domain-specific tokens, lowering acceptance in speculative decoding.  
Other work on large vocabulary LLM training, such as CCE~\citep{wijmans2024cut}, reduces peak memory by fusing cross-entropy directly into the kernel, avoiding explicit logit materialization.  
Our approach is motivated by research in \emph{extreme classification}, where the goal is to accelerate learning and inference over output spaces with millions of labels. Methods such as LightXML~\citep{jiang2021lightxml}, CascadeXML~\citep{kharbanda2022cascadexml}, and related meta classifier based strategies~\citep{zhang2021fast, kharbanda2023inceptionxml} adopt coarse-to-fine or cluster-based mechanisms that limit computation to a small subset of relevant labels per input. These ideas directly inspire \textsc{DynaSpec}: rather than evaluating the drafter across the entire vocabulary or a static shortlist, we employ a lightweight meta-classifier to select a small number of context-relevant clusters, retaining the coverage of rare tokens while reducing compute.

\section{Computational bottleneck of speculative decoding in Large-vocabulary LMs} 
Under standard autoregressive decoding, let \(\tau_T\) denote the per-token latency of the target model.
With speculative decoding, let \(\tau_D\) be the draft model's per-token generation time, and \(\tau_V(\gamma)\) be the time for the target to verify \(\gamma\) the proposed tokens.
Given \(\alpha\in[0,1]\) as the expected acceptance rate and \(\gamma\in\mathbb{N}\) the speculation length, the expected number of tokens produced per verification step is (per equation (1) in \citet{leviathan2023fast}) :
\begin{equation}
\Omega(\alpha,\gamma)
= \mathbb{E}[\#\text{generated tokens}]
= \frac{1-\alpha^{\gamma+1}}{1-\alpha}.
\end{equation}
The average per-token latency under speculative decoding is therefore
\begin{equation*}
\tau_{\mathrm{SD}}
= \frac{\gamma \cdot \tau_D + \tau_V(\gamma)}{\Omega(\alpha,\gamma)}.
\end{equation*}
Hence, the speedup relative to standard autoregressive decoding is
\begin{equation}
\label{eq:speedup}
\mathrm{Speedup}
= \frac{\tau_T}{\tau_{\mathrm{SD}}}
= \frac{\tau_T \cdot \Omega(\alpha,\gamma)}{\gamma \cdot \tau_D + \tau_V(\gamma)}.
\end{equation}

As shown in Equation~\ref{eq:speedup} the speedup of speculative decoding is determined by the verification time $\tau_V(\gamma)$, the draft cost $\tau_D$, and the expected generation length $\Omega(\alpha,\gamma)$. 
Since the target LMs main computation during inference is dominated by multiple stacked attention and MLP layers, the terms $\tau_T$ and $\tau_V(\gamma)$ remain relatively unaffected by increasing vocabularies.
On the other hand, the \emph{drafting} time per-token decomposes into a single transformer layer (attention and MLP) with core term $\Theta(L_D d^2 + L_D S d)$ and a vocabulary-head term $\Theta(|V|d)$. As $V$ increases, the $\Theta(|V|d)$ component can overtake the core, inflating $\tau_D$ and---because it is multiplied by $\gamma$ in the speedup denominator---either reducing the achievable speedup or forcing a smaller $\gamma$. 

Restricting the draft to a smaller subset of the vocabulary, a crude approach taken in FR-Spec \cite{zhao-etal-2025-fr} and Vocab-Trim \cite{goel2025vocabtrim},  effectively reduces the \(\Theta(|V|d)\) term and lowers \(\tau_D\). 
Furthermore, if this subset is not context-aware, the acceptance rate \(\alpha\) may also drop, shrinking \(\Omega(\alpha,\gamma)\) and hence offsetting the latency gains of vocabulary truncation. 

\section{\textsc{DynaSpec} - Context-aware Dynamic Sampling} 

In a nutshell, \textsc{DynaSpec} is a context-aware dynamic shortlisting method for speculative decoding with large-vocabulary LLMs, targeting the drafter’s output-head bottleneck that scales as $\mathcal{O}(|V|d)$. 
Instead of drafting over the full vocabulary (or a brittle, frequency-ranked static shortlist), \textsc{DynaSpec} uses a lightweight router/meta-classifier to select a small set of coarse token clusters for the current context, and the drafter evaluates logits only on the union of tokens in those clusters. Crucially, the target model still verifies over the full vocabulary, so decoding remains exact, while drafting becomes cheaper and better preserves acceptance on rare/domain tokens than fixed shortlists. Systems-wise, the router runs in parallel with draft encoding and \textsc{DynaSpec} uses a position-aware cluster budget (larger early, smaller later) with a fused indexed-head kernel to reduce overhead where it matters most.

Before presenting the training pipeline of \textsc{DynaSpec}, we setup the notation and outline (i) \textsc{Eagle} ~\citep{li2024eagle,li2024eagle2} and (ii) \textsc{FR-Spec} \cite{zhao-etal-2025-fr}, which are most relevant to our work. 

\subsection{Notation and preliminaries} \noindent\textbf{Target LLM.} Let $T$ represent the target LLM over vocabulary of tokens' vocabulary $V$. 
Given $d$-dimensional token embeddings $\mathbf{E}$, for an input sequence $x_{1:n}$, the embedding layer produces
\[Z_0 = \mathbf{E}(x_{1:n}) \in \mathbb{R}^{n\times d}.
\]
Applying $\ell$ transformer blocks $\mathcal{F}_1,\ldots,\mathcal{F}_\ell$ yields
\[
H \;=\; (\mathcal{F}_\ell \circ \cdots \circ \mathcal{F}_1)(Z_0) \in \mathbb{R}^{n\times d}.
\]
An LM head with weights $W_{\mathrm{LM}} \in \mathbb{R}^{d\times |V|}$ maps to logits $O = HW_{\mathrm{LM}}$, and the token probabilities are
\[
p_{T}(y_i \mid x_{\le i}) \;=\; \operatorname{softmax}(O_i).
\]
\noindent\textbf{Full Vocabulary Drafter.} Speculative decoding methods (more specifically \textsc{Eagle}~\citep{li2024eagle,li2024eagle2} series of models) train a lightweight draft model $D$ (one transformer block, same latent width $d$) to approximate $T$ at a lower cost.
The drafter reuses and freezes the embedding $\mathbf{E}$ and LM head $W_{\mathrm{LM}}$ from $T$, and only its single transformer block is trained. The output distribution $q(.)$ over the tokens is computed as :
$\tilde{H} \;=\; D\!\big(\mathbf{E}(x_{1:n})\big)$ and  $q_{\text{Eagle}} \;=\; \operatorname{softmax}(\tilde{H}W_{\mathrm{LM}})$.

\noindent\textbf{Static Sub-vocabulary Drafter.} \textsc{Fr-Spec}~\citep{zhao-etal-2025-fr} uses a frequency-ranked drafting with vocabulary $V_{\text{high}}\!\subset\!V$ the subset of high-frequency tokens (from corpus statistics). Concretely, select the corresponding LM-head columns.
\[
\widetilde{W}\in\mathbb{R}^{d\times|V_{\text{high}}|},\quad
\widetilde{W}[:,j]=W_{\mathrm{LM}}[:,\,V_{\text{high}}[j]],
\]
and compute as $q_{\text{Fr-Spec}}=\operatorname{softmax}\!\big(\tilde{H}\,\widetilde{W}\big)$.

\subsection{Context-aware dynamic sub-vocabulary selection}

\paragraph{Vocabulary partition.} We partition the vocabulary $V$ into $M$ disjoint token clusters i.e. $V = \biguplus_{m=1}^{M} C_m$.  
These clusters are formed offline by spherical $k$-means on column-normalized LM-head weights which is given by : \\ $\{W_{LM}[:,v]/\|W_{LM}[:,v]\|_2\}_{v\in V}$. Since some tokens in the vocabulary share similar semantics, we can cluster them and use a meta-classifier to efficiently select a relevant subset.
We \emph{do not} enforce balance (i.e., $|C_m|$ may vary), which empirically preserves recall for semantically dense regions.

\paragraph{Low-cost router.}
Given the token embedding $\mathbf{E}(x_{t})$ and the drafter’s previous hidden features $\tilde{H}_{t-1}$, a lightweight MLP $r_\theta:\mathbb{R}^{d_r}\!\to\!\mathbb{R}^M$ produces scores $s =r_\theta([\mathbf{E}(x_{t}), \tilde{H}_{t-1}])$ over clusters.  
Given these scores, we select
$\mathcal{K}(c,t)=\TopK_{k_{\mathrm{c}}(t)}(s)$ by a top-$k$ operation, and construct the
shortlist $\mathcal{V}_S(c,t)=\biguplus_{m\in\mathcal{K}(c,t)} C_m$. At each draft step, $x_t$ may consist of a single token or multiple tokens, depending on the beam size at step $t$. When $x_t$ contains multiple tokens, we take the union of the clusters predicted for each token.

\begin{figure}[tb]
    \centering
\includegraphics[width=0.75\columnwidth]{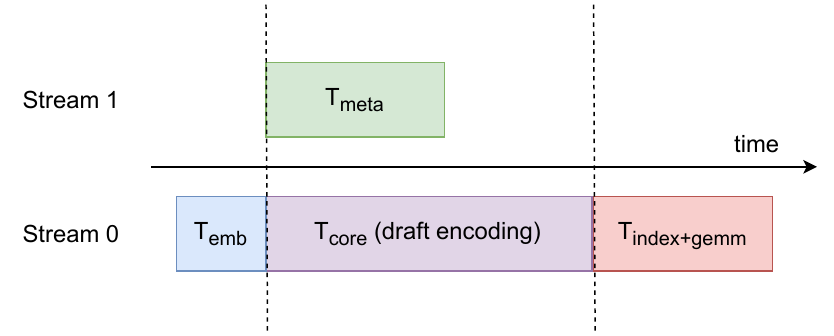}
    \caption{Time breakdown of the drafting process of
 \textsc{DynaSpec}. $T_{\text{emb}}$ is the embedding time, $T_{\text{core}}$ is the encoder layer time, $T_{\text{meta}}$ is the time related to the meta classifier and $T_{\text{index+gemm}}$ is the indexed matmul time. }
    \label{fig:timing}
\end{figure}

Instead of a fixed top-$k$ policy, we use a \emph{position-aware} budget $k_{\mathrm{c}}(t)\in\{k_{\min},\ldots,k_{\max}\}$ that allocates more clusters to early tokens and fewer to later ones. Concretely, with
$t=0$ denoting the first drafted token, we assign a larger initial budget and then decay as
\begin{equation}
\label{eq:pos_schedule}
k_{\mathrm{c}}(t)
=
\begin{cases}
k_{0}, & t = 0,\\[4pt]
k_{1}, & t = 1,\\[4pt]
\left\lfloor \dfrac{k_{1} + t}{2t} \right\rfloor, & t \ge 2,
\end{cases}
\end{equation}

where $k_{0} > k_{1}$ and both are hyperparameters controlling the initial cluster budgets. 
As shown in Figure \ref{fig:timing}, the router runs on a separate parallel CUDA stream and completes while the drafter’s attention/MLP is executing, making its wall-clock overhead negligible. This is because the draft model is typically very lightweight and cannot fully utilize the GPU, allowing the meta-selection to run in parallel on a separate stream.

\paragraph{Training the router.}
We cast training the router $r_\theta$ as a multi-label classifier over $M$ clusters with \emph{hard} labels. 
At each time step $t$,
the router observes the current context features (we use the embedding of the current token and the previous drafter hidden state) and outputs a score vector :
\begin{displaymath}
    s_t = r_{\theta}([\textbf{E}(x_t), \tilde{H}_{t-1]}]) \in \mathbb{R}^M, \hspace{0.1in} \hat{y}_{t,m} = \sigma (s_{t,m}) \in (0,1)
\end{displaymath}
where $M$ is the number of clusters and $\sigma$ is the sigmoid. The router is implemented as a two-layer feed-forward neural network with ReLU activation. Given an input feature vector, the network projects it into a hidden representation and then maps it to cluster logits through a linear layer.

To construct the training labels, we run the \textsc{Eagle}-2 pipeline on a large set of prompts (e.g., ShareGPT and UltraChat), and at each step $t$, we record : (i) the drafter hidden state $\tilde{H}_{t-1}$, (ii) the current token embedding $\textbf{E}(x_t)$, and (iii) a set of plausible next token $S_t^+ \subset V$ (e.g., the top-$L$ tokens under the full-vocabulary drafter at that step). Each token $v \in V$ belongs to exactly one cluster $C_m$. We therefore induce a multi-hot target over clusters :
\begin{displaymath}
    y_{t,m} = \mathbb{I}[C_m \cap S_t^+ \neq \emptyset] \in \{0,1\}.
\end{displaymath}
Intuitively, a cluster is labeled positive if it contains at least one high-probability token for that context. 
We optimize binary cross-entropy (independent logistic losses) over clusters:

\[
\mathcal{L}_{\mathrm{r}}
\hspace{-0.05in}\;=\; \hspace{-0.05in}
\mathbb{E}\!\left[
\sum_{m=1}^{M} \hspace{-0.05in}
\Big(\hspace{-0.05in}-\,y_{t,m}\log \hat{y}_{t,m} \hspace{-0.05in} - (1-y_{t,m})\log\big(1-\hat{y}_{t,m}\big)\Big)
\right]
\]

It must be noted that even though the next token prediction is a multi-class problem, the \textsc{Dynaspec}'s router is trained as multi-label classifier. 
This is because at time step $t$, the supervision signal is a set of tokens $S_t^+$, which land in multiple clusters because they include synonyms, formatting variants, punctuation, code tokens, domain terms, etc. 
Therefore, the induced label set $Y_t = \{m : C_m \cap S_t^+ \neq \emptyset\}$ usually has $|Y_t| > 1$, making it a multi-label problem.

\subsection{Dynamic Drafting}
\label{sec:dynamic-drafting}
\begin{figure*}[t]
    \centering
    \includegraphics[width=0.9\textwidth]{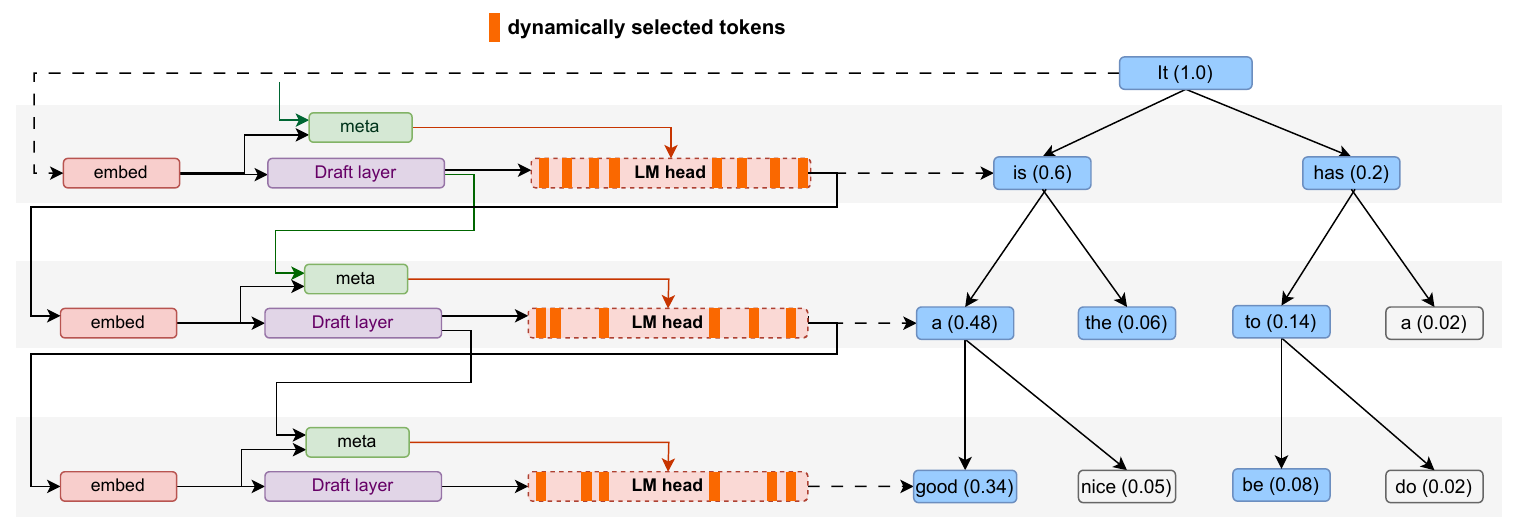}
    \caption{Draft process of DynaSpec. The backbone structure is based on EAGLE where the meta classifier selects the top-k tokens dynamically based on the context.}
    \label{fig:placeholder}
\end{figure*}

The drafting process for a single drafting step is depicted in Figure \ref{fig:placeholder}. 
Given a context $c$ at position $t$, the router first produces cluster scores and selects $\mathcal{K}(c,t)$, which induces the shortlist $\mathcal{V}_S(c,t)$. In parallel, the drafter transformer computes the step hidden state $\tilde{H}_t$. We execute routing on a separate CUDA stream so that router latency overlaps with the drafter’s attention/MLP computation; in practice $T_{meta} \ll T_{core}$ (c.f. \ref{fig:timing}), so routing is typically hidden by the draft forward pass. 

After synchronizing the two streams, we compute drafter logits using a gathered LM head over the shortlisted tokens only. Let $B(c,t) = |\mathcal{V}_S(c,t)|$. Instead of the full projection $\tilde{H}_tW_{\text{LM}}$ with cost $\mathcal{O}(|V|d)$, we gather the $B(c,t)$ corresponding columns of $W_{\text{LM}}$ and perform a single GEMM of cost $\mathcal{O}(B(c,t)d)$ where $B(c,t) \ll |V|$. This yields the following per-step latency approximation:

\[
T_D(c,t)\ \approx\ T_{\text{emb}} +  \max\!\big\{T_{\text{core}},\,T_{\text{meta}}\big\}\ +T_{\text{index+gemm}}\!\big(B(c,t)\big)
\]
where $\max(.,.)$ term reflects the stream overlap between routing and the drafter core. 

Since $T_{\text{index}+ \text{gemm}}$ grows with $B(c,t)$, we use a position-aware cluster budget $k_{\mathrm{c}}(t)$ decay with $t$, allocating larger shortlists early (to protect acceptance) and smaller shortlists later (to reduce compute).

In our implementation, \emph{index selection and the head GEMM are fused} into a single custom CUDA kernel to avoid intermediate buffers and redundant global-memory traffic: the kernel performs warp-level column gathers with asynchronous, double-buffered shared-memory tiling, computes on Tensor Cores, and uses a persistent L2 access-policy window to keep activations cached. Verification is unchanged from standard speculative decoding.

In the next section, we present a general theoretical analysis of speculative decoding when the draft model is restricted to a vocabulary subset. Apart from the experimental comparisons on benchmark datasets later in the paper, the theoretical framework can also be used to analytically compare concrete variants such as \textsc{DynaSpec} and \textsc{Fr-Spec}.

\section{Theoretical Analysis} 
For vocabulary $V$, at a decoding step with context $c$, let the target distribution be denoted as $p_c(x) \coloneqq p(x \mid c), x \in V$, and the full vocabulary draft distribution be denoted as $q_c(x) \coloneqq q(x \mid c) $. Note that this is only conceptual in our setting and corresponds to \textsc{Eagle}-style drafter if it had a full head. \textsc{DynaSpec}'s router chooses a context-dependent shortlist $S(c) \subset V$. 
The router selects a set of cluster indices $\mathcal{K}(c) \subset [M], |\mathcal{K}(c)| = k$, and tokens are shortlisted in as $S(c) = \cup_{m \in \mathcal{K}(c)} C_m$, also in context-dependent manner. The restricted and renormalized draft distribution on $S$ is given by (denoted as $q_{S,c}(x)$) :
\begin{displaymath}
  \frac{q_c(x) \mathbb{I}[ x \in S(c)]}{q_c(S(c))} \eqqcolon q_{S,c}(x)
\end{displaymath}
Define a tail distribution $\omega_c(\cdot)$ supported on $V \setminus S(c)$. 
Then the actual proposal distribution is given by :
 \begin{displaymath}
     \tilde{q}_c (x) \coloneqq (1-\beta) q_{S,c}(x) + \beta \omega_c(x)
 \end{displaymath}
with $\beta \in (0,1)$ but with a small value. 
The tail-distribution $\omega_c(x)$ makes the proposal well-defined (full support) and lets us reason about correctness and acceptance. Also, it cleanly separates ``shortlist quality'' (through $q_S$) from the safety tail $(\beta, \omega)$, which is important for modelling purposes. Note that, $\omega_c(\cdot)$ can be a uniform distribution over the complement or a cheap approximate tail distribution.  
In practice, even if the implementation does not explicitly sample from a tail, we can set $\beta$ to an abstract ``escape'' probability that models either (a) fallbacks or (b) the effect of verifying and then resampling at rejection, making the theory robust to these details.

For one speculative step (one position), let $p_c(x)$ be the target next-token distribution given context $c$, and $\tilde{q}_c(x)$ be the effective draft proposal distribution for that step. 
The overlap between the two is denoted by $A(c)$ : 
\begin{equation}
\label{eq:overlap}
    A(c) \coloneqq \sum_{x \in V} \min(p_c(x), \tilde{q}_c(x)) = 1 - \text{TV}(p_c(x), \tilde{q}_c(x))
\end{equation}
where $\text{TV}(.,.)$ is the total variation distance between distributions given by : 
\begin{equation}
\text{TV}(p, q) \coloneqq \frac{1}{2}\sum_{x \in V} |p(x) - q(x)|
\end{equation}

Write shorthand for $S(c)$ as : $S(c) = S$, and denoting by $p_c(S) \coloneqq \sum_{x \in S(c)}p_c(x)$, the target mass retained by the router shortlist $S$. Let $p_c(\cdot \mid S)$ and $q_c(\cdot \mid S)$ are the target and drafter distribution conditioned on the restriction to the shortlist $S$, i.e. $p_c(x \mid S) = p_c(x)/p_c(S)$ if $x \in S$, and 0 otherwise. Similarly, for $q_c(\cdot \mid S)$, start with the drafter's full-vocabulary distribution $q_c(x), x \in V$(conceptual in our scenario), and compute its mass on $S$, i.e., $q_c(S) \coloneqq \sum_{x \in S(c)}q_c(x)$, then $q_c(x \mid S) = q_c(x)/q_c(S)$ if $x \in S$, and 0 otherwise. 
\begin{thm} [Distribution overlap lower bound]
    For any context $c$,
    \begin{displaymath}
        A(c) \geq (1-\beta) \left( p_c(S) - \textnormal{TV} ( p_c(\cdot \mid S), q_c(\cdot \mid S) )\right). 
    \end{displaymath} 
\end{thm}\label{thm:tv}
    
The proof is given appendix \ref{sec:proof}. 
Essentially, the theorem above suggests that two issues can hurt acceptance. Firstly, coverage problem which is captured by $p_c(S)$, it is small when most target mass lies outside $S$. Secondly, even if $p_c(S)$ is large, the drafter might put probability on the wrong token within $S$, which is the quantity captured by $\text{TV}(p_c(\cdot \mid S), q_c(\cdot \mid S))$. 
Concretely, using the corresponding shortlists computed by \textsc{Dynaspec} and \textsc{FR-Spec}, i.e., $S_{\text{Dyn}}$ and $S_{\text{Fr}}$ respectively, one can compute the key quantities $p_c(S)$ and $\textnormal{TV} ( p_c(\cdot \mid S), q_c(\cdot \mid S) )$ in the above theorem \ref{thm:tv} and compare the overlap lower bounds. 
These quantities are computed for the HumanEval (Code) and compared for the \textsc{Dynaspec} and \textsc{FR-Spec} in Table \ref{tab:results-theory} (in Apendix A.2 along with the proof of the Theorem).

\begin{cor}[Routing regret]
    Let $S^*(c)$  be an oracle shortlist (e.g., the best union-of-$k$-clusters set under $p_c$), define the routing regret as 
    $\epsilon(c) \coloneqq p_c(S^*(c)) - p_c(S(c)) \geq 0$, then 
    \begin{displaymath}
         A(c) \geq (1-\beta) \left( p_c(S^*(c)) - \epsilon (c) - \textnormal{TV} ( p_c(\cdot \mid S), q_c(\cdot \mid S) )\right).
    \end{displaymath}
\end{cor}

Note that \textsc{FR-Spec} has a static regret $\epsilon_{Fr}(c) = p_c(S^*(c)) - p_c(S_{FR})$ while the corresponding quantity $\epsilon_{Dyn}(c) = p_c(S^*(c)) - p_c(S_{Dyn}(c))$ is dynamic for \textsc{Dynaspec}. 
In particular, for an accurate router, then $\epsilon_{Dyn}(c)$ would be smaller on hard contexts such as code and rare symbols.

\begin{thm} [Expected accepted length] \label{thm:seq_length}
   Consider a block proposal of length $b$, and let the contexts along the draft be $c_1, c_2, \ldots, c_b$ such that each depends on previously proposed tokens. Let $A_i \coloneqq A(c_i)$ be the overlap at step $i$. Let $L \in \{0,1, \ldots,b \}$ be the number of consecutively accepted draft tokens before the first rejection. If each step uses a proposal $\tilde{q}_{c_i}$ and the acceptance test is the standard speculative accept/reject, so that the probability of accepting token $i$ given it was reached is atleast $A_i$, then  
\begin{equation}
\label{eq:sequence}
     \mathbb{E}[L] \geq \sum_{i=1}^b \prod_{j=1}^i A_j.
\end{equation}

\end{thm}
The proof is given appendix \ref{sec:proof}. Plugging in $A_j = A_{\text{FR}}(c_j)$ or $A_{\text{Dyn}}(c_j)$ gives a method to compare the mean accepted lengths. 
The multiplicative form for the expected accepted length as highlighted in Theorem \ref{thm:seq_length} shows that by having a larger pool of tokens to choose from, in the earlier steps of the drafting process, \textsc{Dynaspec} can outperform on the contexts where static $S_{Fr}$ misses mass. 
The above analysis also provides an analytical justification to the choice of having a position aware budgeting strategy (Equation \eqref{eq:pos_schedule}) used in the router design.

\section{Experimental Results}

\subsection{Experimental Settings} We adhere to the experimental setup and evaluation of the \textsc{Eagle}~\citep{li2024eagle,li2024eagle2} family and \textsc{FR-Spec}~\citep{zhao-etal-2025-fr}. We implement the \textsc{DynaSpec} in the \textsc{FR-Spec}~\footnote{\url{https://github.com/thunlp/FR-Spec}} framework. All components are implemented in native C++ and CUDA with FR-Spec as the baseline, since Python call overhead would otherwise negate the latency gains from draft-time acceleration.

\textbf{Datasets and Models.} We benchmark DynaSpec over a diverse mix of seven tasks. Spec-Bench~\citep{xia2024unlocking} supplies six of them---Machine Translation (WMT14 DE to EN)~\citep{bojar2014findings}, Multi-turn Conversation (MT-Bench)~\citep{zheng2023eric}, RAG and QA from Natural Questions~\citep{kwiatkowski2019natural}, Math from GSM8K~\citep{cobbe2021training}, and Summarization (CNN/DailyMail)~\citep{nallapati2016abstractive}---each evaluated on 80 prompts. The seventh task is Code, for which we employ HumanEval~\citep{chen2021evaluating} with 164 problems. We bound generations to $1024$ tokens for Spec-Bench subtasks and 512 tokens for HumanEval. We used Llama series of models, specifically Llama-$3$-8B-Instruct, which uses a vocabulary size of $128K$ for our experiments and Qwen-2-7B.

\textbf{Evaluation metrics and Hardware Settings.} We evaluate using  metrics: 
\#Mean Acceptance Length, the average number of tokens committed per draft–verification cycle. \#Mean Acceptance Length is hardware-agnostic but does not capture the overhead of drafting. Our primary baseline is \textsc{FR-Spec}~\citep{zhao-etal-2025-fr}, a static subset–selection approach built on \textsc{Eagle}-2~\citep{li2024eagle2}; in contrast, \textsc{DynaSpec} performs dynamic subset selection atop \textsc{Eagle}-2~\citep{li2024eagle2}.  
All experiments were run on a single NVIDIA A6000 GPU.

\begin{table*}[t]
\caption{Mean accepted tokens for \textbf{Llama-3-8B} and \textbf{Qwen-2-7B} across seven datasets. vocab size indicates the shortlisted number of tokens. Full Vocab  corresponds to the \textsc{EAGLE-2} baseline; \textsc{FR-Spec} uses 32k. For \textsc{DynaSpec}, shortlist size varies by dataset.}
\centering
\scriptsize
\begin{adjustbox}{width=\textwidth}
\begin{tabular}{l
                l
                S[table-format=1.2]
                S[table-format=1.2]
                S[table-format=1.2]
                S[table-format=1.2]
                S[table-format=1.2]
                S[table-format=1.2]
                S[table-format=1.2]
                S[table-format=1.2]}
\toprule
\multicolumn{2}{c}{\textbf{Model / Config}}
  & \multicolumn{1}{c}{\textbf{MT}}
  & \multicolumn{1}{c}{\textbf{Conv.}}
  & \multicolumn{1}{c}{\textbf{RAG}}
  & \multicolumn{1}{c}{\textbf{Math}}
  & \multicolumn{1}{c}{\textbf{QA}}
  & \multicolumn{1}{c}{\textbf{Summ.}}
  & \multicolumn{1}{c}{\textbf{Code}}
  & \multicolumn{1}{c}{\textbf{Average}} \\
\midrule

\multirow{9}{*}{\textbf{Llama-3-8B}}
  & Full Vocab (Upper bound)
  & 3.65 & 4.13 & 4.04 & 4.32 & 3.46 & 3.69 & 3.92 & 3.89 \\
  & {\color{gray}\scriptsize \textit{vocab size=128k}}
  & {\color{gray}\scriptsize --} & {\color{gray}\scriptsize --} & {\color{gray}\scriptsize --} & {\color{gray}\scriptsize --}
  & {\color{gray}\scriptsize --} & {\color{gray}\scriptsize --} & {\color{gray}\scriptsize --} & {\color{gray}\scriptsize --} \\
\addlinespace[1pt]
  & FR-Spec
  & 3.39 & 3.88 & 3.86 & 4.16 & 3.33 & 3.52 & 3.31 & 3.64 \\
  & {\color{gray}\scriptsize \textit{vocab size=32k}}
  & {\color{gray}\scriptsize --} & {\color{gray}\scriptsize --} & {\color{gray}\scriptsize --} & {\color{gray}\scriptsize --}
  & {\color{gray}\scriptsize --} & {\color{gray}\scriptsize --} & {\color{gray}\scriptsize --} & {\color{gray}\scriptsize --} \\
\addlinespace[1pt]
  & DynaSpec
  & 3.59 & 4.10 & 3.98 & 4.27 & 3.41 & 3.58 & 3.86 & 3.83 \\
  & {\color{gray}\scriptsize \textit{vocab size$\approx$28k }}
  & {\color{gray}\scriptsize -} & {\color{gray}\scriptsize -} & {\color{gray}\scriptsize -}
  & {\color{gray}\scriptsize -} & {\color{gray}\scriptsize -} & {\color{gray}\scriptsize -}
  & {\color{gray}\scriptsize -} & {\color{gray}\scriptsize -} \\
  \midrule

\multirow{9}{*}{\textbf{Qwen-2-7B}}
  & Full Vocab (Upper bound)
  &  2.92& 3.79 & 3.43 & 4.34 & 3.07 & 3.43 &4.17 & 3.59 \\
  & {\color{gray}\scriptsize \textit{vocab size=152k}}
  & {\color{gray}\scriptsize --} & {\color{gray}\scriptsize --} & {\color{gray}\scriptsize --} & {\color{gray}\scriptsize --}
  & {\color{gray}\scriptsize --} & {\color{gray}\scriptsize --} & {\color{gray}\scriptsize --} & {\color{gray}\scriptsize --} \\
\addlinespace[1pt]
  & FR-Spec
  & 2.84 & 3.52 & 3.30 & 4.17 & 2.95 & 3.32 & 3.52 &3.37  \\
  & {\color{gray}\scriptsize \textit{vocab size=32k}}
  & {\color{gray}\scriptsize --} & {\color{gray}\scriptsize --} & {\color{gray}\scriptsize --} & {\color{gray}\scriptsize --}
  & {\color{gray}\scriptsize --} & {\color{gray}\scriptsize --} & {\color{gray}\scriptsize --} & {\color{gray}\scriptsize --} \\
\addlinespace[1pt]
  & DynaSpec
  & 2.88 & 3.76 & 3.39 & 4.28 & 3.02 & 3.33 & 4.08 & 3.53 \\
  & {\color{gray}\scriptsize \textit{vocab size$\approx$20k}}
  & {\color{gray}\scriptsize --} & {\color{gray}\scriptsize --} & {\color{gray}\scriptsize --}
  & {\color{gray}\scriptsize --} & {\color{gray}\scriptsize --} & {\color{gray}\scriptsize --}
  & {\color{gray}\scriptsize --} & {\color{gray}\scriptsize --} \\
\bottomrule
\end{tabular}
\end{adjustbox}
\label{tab:main}
\end{table*}

\begin{figure}[h]
    \centering
    \includegraphics[width=0.40\textwidth]{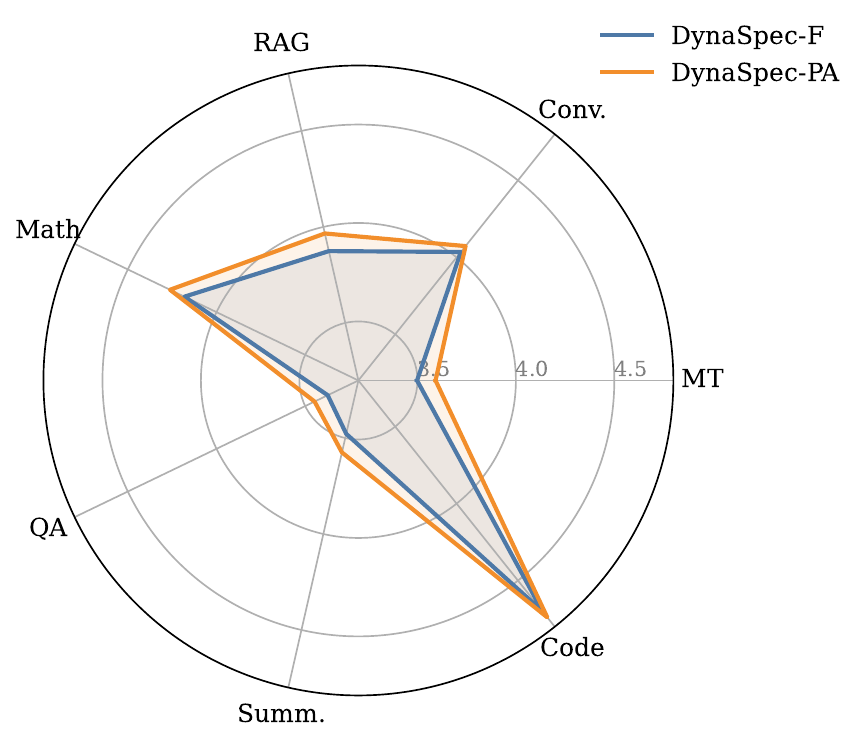}
    \caption{Mean accepted tokens comparison for fixed top-$k$ clusters represented by DynaSpec-F vs. Position-aware top-$k$ clusters represented by DynaSpec-PA across seven datasets using Llama-3-8B model. The mean vocabulary sizes for DynaSpec-PA and DynaSpec-F are 28.7k and 29.2k, respectively. }
    \label{fig:fix_vs_dynanmic}
\end{figure}

\begin{table*}[t]
\centering
\caption{Average tokens/sec for Llama-$3$-8B-Instruct on seven datasets. Numbers in parentheses under \textbf{Average} indicate the ratio vs. the Target baseline.}
\scriptsize
\resizebox{\textwidth}{!}{
\begin{tabular}{lcccccccc}
\toprule
\textbf{Method} & \textbf{MT.} & \textbf{Conv.} & \textbf{RAG} & \textbf{Math} & \textbf{QA} & \textbf{Summ.} & \textbf{Code} & \textbf{Average} \\
\midrule
Target LLM (Tokens/s)        & 43.02 & 42.83 & 39.50 & 43.17 & 43.06 & 41.08 & 42.60 & 42.18 (1.00$\times$) \\
\midrule
EAGLE-2 (full vocab)     & 79.37 & 91.34 & 75.09 & 94.26 & 74.07 & 74.57 & 81.59 & 81.47 (1.93$\times$) \\
FR-Spec-32k        & 87.56 & 101.85 & 81.81 & 107.91 & 84.61 & 83.46 & 81.60 & 89.83 (2.13$\times$) \\
DynaSpec-27k            & \textbf{92.97} & \textbf{107.55} & \textbf{85.55} & \textbf{110.96} & \textbf{87.11} & \textbf{85.71} & \textbf{95.01} & \textbf{94.98} (\textbf{2.25$\times$}) \\
\bottomrule
\end{tabular}
}
\label{tab:speed}
\end{table*}

\subsection{Mean Acceptance Length}
We evaluate \textit{mean accepted length}--the average number of draft tokens committed per draft verification cycle-- across seven benchmarks. 
Table~\ref{tab:main} reports results for Llama-3-8B-Instruct and Qwen-2-7B under three settings: (i) full-vocabulary \textsc{Eagle}-2 (upper bound), (ii) \textsc{FR-Spec} with a fixed 32K shortlist, and (iii) \textsc{DynaSpec} with a \textbf{context-dependent} shortlist whose average size is smaller than 32K.

\noindent \textbf{Llama-3-8B.} \textsc{DynaSpec} improves the average accepted length from 3.64 $\rightarrow$ 3.83. tokens/step (+5.2\%) while using a smaller mean shortlist ($\sim 28K$ vs \textsc{FR-Spec}'s fixed 32K). Relative to full-vocabulary decoding (3.89), \textsc{DynaSpec} recovers $\sim 98.4\%$ of the upper bound, whereas \textsc{FR-Spec} recovers 93.6\%. Gains are consistent across tasks, while Code setting shows the largest gap: static frequency truncation drops substantially below full-vocab (3.92), while \textsc{DynaSpec} nearly matches it (3.86), indicating that context-aware routing better preserves probability mass on rare identifiers, symbols, and formatting tokens that are systematically under-represented in frequency-ranked lists. 

\noindent \textbf{Qwen-2-7B.} We observe the same pattern: DynaSpec improves the average accepted length from 3.37 $\rightarrow$ 3.53. while using an even smaller average shortlist ($\sim$ 20K vs 32K). Compared to full-vocabulary decoding (3.59), this corresponds to $\sim$ 98.3\% recovery for DynaSpec vs. $\sim$93.9\% for FR-Spec. Per-task improvements mirror the Llama results (notably Conversation 3.52 $\rightarrow$ 3.76 and Code 3.52 $\rightarrow$ 4.08, suggesting the benefit of dynamic, context-conditioned supports is robust across model families and vocabulary sizes. 

Overall, these results support the central claim: dynamic, context-aware shortlisting can match (or closely approach) full-vocabulary acceptance while operating at substantially smaller effective vocabularies than fixed shortlists, which is exactly the regime needed to translate draft-head savings into end-to-end speculative decoding speedups, as described in the next section.

\vspace{-0.1in}
\subsection{Decoding Speed}
We measure end-to-end throughput (tokens/sec) for Llama-3-8B-Instruct on a single NVIDIA A6000, using a CUDA/C++ implementation to avoid Python overhead. We compare Target-only decoding, full-vocabulary \textsc{Eagle}-2, \textsc{FR-Spec}-32k (static shortlist), and \textsc{DynaSpec}-27k (dynamic routed shortlist). 

As shown in Table 2, \textsc{DynaSpec} achieves 94.98 tokens/s on average (2.25$\times$ over Target-LLM only), improving over both EAGLE-2 full-vocab (81.47 tokens/s; 1.93$\times$) and \textsc{FR-Spec}-32k (89.83 tokens/s; 2.13$\times$). 
Per-task, DynaSpec is faster than FR-Spec on all seven datasets.
The largest gain is on Code (+13.41 tokens/s), consistent with context-aware routing improving tail-token coverage while still reducing draft-head compute. Overall, DynaSpec improves average throughput by +5.15 tokens/s over FR-Spec ( +5.7\%).

\subsection{Position-Aware Vocabulary Shortlisting}

\newcommand{\best}[1]{\textbf{#1}}

For the position-aware strategy, the results are presented in Figure~\ref{fig:fix_vs_dynanmic}. Using a position-aware subset size outperforms a fixed size across all draft steps, as the first few tokens have a disproportionate impact on acceptance and throughput, an observation consistent with prior findings in \textsc{Gumiho}~\cite{li2025gumiho}. Allocating a larger cluster budget to early tokens preserves acceptance where it matters most, while decaying the shortlist later reduces indexed-GEMM cost without affecting verification. As shown in Figure~\ref{fig:fix_vs_dynanmic}, this adaptive schedule yields higher mean accepted lengths at comparable or smaller shortlist sizes. Table~\ref{tab:position-aware-frspec} (in Appendix) further reports results for combining the position-aware schedule with FR-Spec, which likewise improves mean acceptance and reduces average shortlist size. Overall, these results confirm that position awareness complements context-aware routing by focusing computation on the most influential draft steps and amortizing overhead in later ones.

\vspace{-0.1in}

\section{Conclusion}
We address a central bottleneck in speculative decoding for large-vocabulary LLMs---the drafter’s $\mathcal{O}(|V|d)$ output head. \textsc{DynaSpec} replaces static shortlists with a \emph{context-aware} dynamic head that routes each prefix to a small number of token clusters and evaluates only over their union, while verification remains exact on the full vocabulary. Our analysis shows that context-conditioned supports strictly dominate any fixed subset in expected acceptance, and BCE-with-top-$k$ routing maximizes a lower bound on retained target mass. Systems-wise, we fuse index selection with the head GEMM and run routing in parallel, reducing draft-time cost. Empirically, \textsc{DynaSpec} consistently improves speculative decoding efficiency across standard benchmarks. It also achieves sustains high acceptance quality with substantially smaller candidate sets, while remaining plug-in compatible with EAGLE-style pipelines.

\section*{Impact Statement}
This work improves the computational efficiency of LLM inference by reducing the drafter-side cost in speculative decoding through context-aware dynamic shortlisting, while leaving the target model’s full-vocabulary verification unchanged and therefore not altering the correctness of the decoding procedure. 

Positive impacts include lower latency and potentially reduced energy/compute cost per generated token for deployments that already use speculative decoding. At the same time, efficiency improvements can also lower the marginal cost of generating large volumes of text, which may amplify downstream misuse (e.g., spam or automated content generation) depending on the application and surrounding safeguards. DynaSpec itself is a systems/decoding contribution and does not add new capabilities beyond faster generation; responsible deployment should rely on standard policy, access control, and content-moderation measures at the application level.

\bibliography{references}
\bibliographystyle{plainnat}

\appendix
\section{Appendix}
\subsection{\textsc{Dynaspec} algorithmic procedure}
\begin{algorithm}[H]
\caption{DynaSpec draft steps (single verification cycle)}
\label{alg:dyna-draft}
\begin{algorithmic}[1]
\REQUIRE prefix sequence $c$, speculation length $\gamma$, target model $T$, router $r_\theta$, drafter $D$, LM head $W_{LM}$, clusters $\{C_m\}$ over vocabulary, first token cluster budget $k_{\mathrm{first}}$, second token cluster budget $k_{\mathrm{second}}$, token budget $k_{\mathrm{t}}$, empty draft list $d$, empty draft list score $\mathrm{d\_scores}$
\FOR{$j=0$ to $\gamma$}
\IF {$j = 0$}
  \STATE \textbf{(stream $S_m$)} $s \leftarrow r_\theta([h_{c[-1]}, emb(x_j)])$;
  \STATE $\mathcal{K}\leftarrow \TopK_{k_{\mathrm{first}}}(s)$;\quad $I \leftarrow \mathrm{indices}\!\big(\cup_{m\in\mathcal{K}} C_m\big)$
  \STATE \textbf{(stream $S_d$)} $\tilde{h} \leftarrow \mathcal{D}\!\big([h_{c}, c[1:] + x_{j}]\big)$
  \STATE $\mathrm{last\_step\_scores} \leftarrow 0$
\ELSE
     \STATE \textbf{if} $j > 1$ \textbf{then} $k_{\mathrm{c}} \leftarrow k_\mathrm{second} // (i*2)$ \textbf{else}  $k_{\mathrm{c}} \leftarrow k_{\mathrm{second}}$
  \STATE \textbf{(stream $S_m$)} $s \leftarrow r_\theta([h_{j-1}, \mathrm{emb}(x_j)])$;\quad $\mathcal{K}\leftarrow \TopK_{k_{\mathrm{c}}}(s)$;\quad $I \leftarrow \mathrm{indices}\!\big(\cup_{m\in\mathcal{K}} C_m\big)$
  \STATE \textbf{(stream $S_d$)} $\tilde{h} \leftarrow \mathcal{D}\!\big([h_{j-1}, x_{j}]\big)$
\ENDIF 
  
  \STATE \textbf{sync} $S_m,S_d$;
  \STATE  $z \leftarrow \mathrm{FUSED\_INDEX\_GEMM}\!\big(\tilde{h},\ W_{LM},\ I\big)$ \Comment{shortlist logits on gathered columns $I$}
  \STATE $p \leftarrow \operatorname{log\_softmax}(z)$;
  \quad $T_j, TopP_j\leftarrow \TopK_{k_{\mathrm{t}}}(p)$; \quad $\tilde{T_j} = \operatorname{remap2realid}(T_j)$
  \STATE $\mathrm{cu\_scores} \leftarrow \operatorname{TopP}_j + \mathrm{last\_step\_scores} $
   \STATE $d \leftarrow d \oplus \tilde{T}_{j}$  \quad $\mathrm{d\_score} \leftarrow  \mathrm{d\_score} \oplus   \mathrm{cu\_scores} $ \Comment{append score and token to the draft list}
  \STATE $\operatorname{TopC}_j, \mathrm{last\_step\_scores} \leftarrow  \operatorname{TopK}_{k_\mathrm{t}}(\mathrm{cu\_scores})$
  \STATE $x_j \leftarrow \tilde{T}_j[\operatorname{TopC}_j]$
  \STATE \textbf{if} $j = 0$ \textbf{then} $h_j \leftarrow \tilde{h}[-1]$ \textbf{else} $h_{j} \leftarrow \tilde{h}[\operatorname{TopC}_j]$
\ENDFOR
\STATE Re-ranking the $d$ based on their corresponding scores $\mathrm{d\_scores}$
\STATE verify once with the target model $T$ over full $\mathcal{V}$ on the top-$k$ subset of $d$; 
\end{algorithmic}
\end{algorithm}

\subsection{Proofs} \label{sec:proof}
\begin{thm} [Distribution overlap lower bound]
    For any context $c$,
    \begin{displaymath}
        A(c) \geq (1-\beta) \left( p_c(S) - \textnormal{TV} ( p_c(\cdot \mid S), q_c(\cdot \mid S) )\right). 
    \end{displaymath} 
\end{thm}
\begin{proof}
    Starting from the overlap identity 
    \begin{displaymath}
        A(c) = \sum_{x \in V } \min(p_c(x), \tilde{q}_c(x)) \geq \sum_{x \in S } \min(p_c(x), \tilde{q}_c(x)).
    \end{displaymath}
    Since, for $x \in S, \tilde{q}_c(x) \geq (1-\beta) q_{S,c}(x) $, then 
    \begin{displaymath}
      A(c) \geq  \sum_{x \in S } \min(p_c(x), \tilde{q}_c(x)) \geq (1-\beta)  \sum_{x \in S } \min(p_c(x), q_{S,c}(x)).
    \end{displaymath}
    Writing $p_c(x) = p_c(S)p_c(x \mid S)$. Also, $q_{S,c}(x) = q_c(x \mid S)$ by definition. Then 
    \begin{displaymath}
        A(c) \geq (1-\beta)  \sum_{x \in S } \min(p_c(S)p_c(x \mid S), q_{c}(x \mid S))  .
    \end{displaymath}
    Also, for each $S$, 
    \begin{displaymath}
        \min(p_c(S)p_c(x \mid S), q_{c}(x \mid S)) \geq p_c(S) \min(p_c(x \mid S), q_{c}(x \mid S)).
    \end{displaymath} 
    Therefore, we have
    \begin{displaymath}
        A(c) \geq (1-\beta) p_c(S) \sum_{x \in S } \min(p_c(x \mid S), q_{c}(x \mid S)).  
    \end{displaymath}
    
    Using the fact that $\sum_{x \in S} \min (a_x, b_x) = 1- \text{TV}(a,b)$ for distributions on $S$. This gives 
    \begin{displaymath}
        A(c) \geq (1-\beta) p_c(S) 
        (1- \text{TV}(p_c(. \mid S), q_c(. \mid S))).
    \end{displaymath}
    Also, since $p_c(S) \leq 1$
    \begin{displaymath}
        p_c(S)(1-\text{TV}(p_c(. \mid S), q_c(. \mid S)) \geq p_c(S) - \text{TV}(p_c(. \mid S), q_c(. \mid S)),
    \end{displaymath} which completes the proof.
\end{proof}
The Table 3 below shows the empirical computation of the quantities of interest in the above lower bound and respective comparison between \textbf{Fr-Spec} and \textsc{Dynaspec}.

\begin{table}[ht]
\centering
\caption{Computation theoretical metrics on the HumanEval (Code) dataset, evaluated over 500 contexts. \textsc{DynaSpec} adopts the 28k vocabulary size.}
\label{tab:results-theory}
\begin{tabular}{lr}
\toprule
\textbf{Metric} & \textbf{Value} \\
\midrule
$p_c(S_{\text{FR-Spec}})$ & 0.8544 \\
$p_c(S_{\text{DynaSpec}})$ & 0.87435 \\
$TV_{\text{FR\_Spec}}$ & 0.88697 \\
$TV_{\text{DynaSpec}}$ & 0.87447 \\
$p_c(S_{\text{FR\_Spec}}) - TV_{\text{FR\_Spec}}$ & $-0.0325$ \\
$p_c(S_{\text{DynaSpec}} - TV_{\text{DynaSpec}}$ & $-0.0001$\\
\bottomrule
\end{tabular}
\end{table}

\begin{thm} [Expected accepted length]
   Consider a block proposal of length $b$, and let the contexts along the draft be $c_1, c_2, \ldots, c_b$ such that each depends on previously proposed tokens. Let $A_i \coloneqq A(c_i)$ be the overlap at step $i$. Let $L \in \{0,1, \ldots,b \}$ be the number of consecutively accepted draft tokens before the first rejection. If each step uses a proposal $\tilde{q}_{c_i}$ and the acceptance test is the standard speculative accept/reject, so that the probability of accepting token $i$ given it was reached is atleast $A_i$, then  
\begin{equation}
\label{eq:sequence}
     \mathbb{E}[L] \geq \sum_{i=1}^b \prod_{j=1}^i A_j.
\end{equation}

\end{thm}

\begin{proof}
Let $E_i$ represent the event that the first $i$ draft tokens are accepted. Then $L\geq i \iff E_i$ occurs. Therefore, 
\begin{displaymath}
    \mathbb{E}[L] = \sum_{i=1}^b Pr(L \geq i) = \sum_{i=1}^b \text{Pr}(E_i).
\end{displaymath}
By chain rule, 
\begin{displaymath}
    \text{Pr}(E_i) = \prod_{j=1}^i \text{Pr}( \text{accept } j \mid E_{j-1}). 
\end{displaymath}
Under the standard speculative accept/reject criterion, a tight lower bound on $\text{Pr}( \text{accept } j \mid E_{j-1})$ is the overlap $A_j$. This gives the desired result.
\end{proof}

\subsection{Position-aware frequency-ranked subset (PA-FR)}
\label{app:pa-fr}
We ask whether \emph{position importance} can also help semi-static, frequency-ranked shortlists. Let $\pi_f:\mathcal{V}\!\to\!\{1,\ldots,|\mathcal{V}|\}$ be the rank order induced by corpus token frequencies (lower is more frequent). Define a position-dependent budget $K_{\mathrm{fr}}(t)$ and construct a shortlist by taking the top-$K_{\mathrm{fr}}(t)$ most frequent tokens:

\[
K_{\mathrm{fr}}(t)
=
\begin{cases}
k_{\max}, & t\in\{0,1\},\\[4pt]
\left\lfloor \dfrac{k_{\max}}{(t+1)} \right\rfloor, & t\ge 2.
\end{cases}
\]
This \emph{position-aware} schedule allocates a larger subset to early tokens, and decays the subset as generation proceeds, reducing head compute without retraining. Compared to a fixed frequency shortlist, PA-FR tends to improve mean accepted length at early steps while lowering average draft-time FLOPs, since the per-step head multiply scales with $K_{\mathrm{fr}}(t)$ rather than a constant $K$.

\begin{table*}[t]
\centering
\scriptsize
\begin{adjustbox}{width=\textwidth}
\begin{tabular}{l
                S[table-format=1.2]
                S[table-format=1.2]
                S[table-format=1.2]
                S[table-format=1.2]
                S[table-format=1.2]
                S[table-format=1.2]
                S[table-format=1.2]
                S[table-format=1.2]}
\toprule
\multicolumn{1}{c}{\textbf{Config}}
  & \multicolumn{1}{c}{\textbf{MT}}
  & \multicolumn{1}{c}{\textbf{Conv.}}
  & \multicolumn{1}{c}{\textbf{RAG}}
  & \multicolumn{1}{c}{\textbf{Math}}
  & \multicolumn{1}{c}{\textbf{QA}}
  & \multicolumn{1}{c}{\textbf{Summ.}}
  & \multicolumn{1}{c}{\textbf{Code}}
  & \multicolumn{1}{c}{\textbf{Avg.}} \\
\midrule
FR-Spec-F
  & 3.38 & 3.87 & 3.85 & 4.16 & 3.32 & 3.51 & 4.11 & 3.74 \\
{\color{gray}\scriptsize \textit{vocab size = 32{,}768}}
  & \multicolumn{8}{c}{} \\
\addlinespace[1pt]
\midrule
FR-Spec-PA
  & \best{3.50} & \best{3.94} & \best{3.86} & \best{4.18} & \best{3.38} & \best{3.58} & \best{4.21} & \best{3.81} \\
{\color{gray}\scriptsize \textit{vocab size = 31{,}739}}
  & \multicolumn{8}{c}{} \\
\bottomrule
\end{tabular}
\end{adjustbox}
\caption{Mean accepted tokens comparison for Fixed top-$k$ tokens represented by FR-Spec-F vs. Position-aware top-$k$ tokens represented by FR-Spec-PA across seven datasets using Llama-3-8B model. Vocab size represents the mean shortlisted vocab size.}
\label{tab:position-aware-frspec}
\end{table*}

\end{document}